\documentclass[letterpaper, 10 pt, conference]{main_content/ieeeconf}
\IEEEoverridecommandlockouts  
\usepackage{graphicx}
\usepackage{epsfig}
\usepackage{times}
\usepackage{mathptmx}
\usepackage{amsmath}
\usepackage{amssymb}
\usepackage{subcaption}
\usepackage{booktabs}
\usepackage{float}
\usepackage{placeins}
\usepackage{tikz}
\usetikzlibrary{positioning, arrows.meta, calc}
\usepackage{fontawesome5}
\usepackage{soul}
\usepackage[percent]{overpic}
\usepackage{colortbl}
\usepackage{nicematrix}
\usepackage{textcomp}
\usepackage{stfloats}
\usepackage{url}
\usepackage{verbatim}
\usepackage{tabularx}
\usepackage{makecell}
\usepackage{multirow}
\usepackage{mwe}
\usepackage{lipsum}
\usepackage{array}
\usepackage{eucal}
\usepackage{pifont}
\usepackage{wasysym}
\usepackage[most]{tcolorbox}
\usepackage[export]{adjustbox}
\usepackage[dvipsnames]{xcolor}
\definecolor{myblue}{RGB}{70,130,200}
\definecolor{lavenderbg}{RGB}{245, 240, 255}
\definecolor{lavenderborder}{RGB}{128, 100, 160}
\usepackage[noadjust]{cite} 
\usepackage[dvipsnames]{xcolor}
\definecolor{myblue}{RGB}{70,130,200}
 
\usepackage{adjustbox}

\title{%
  \LARGE \bf
    Layover or Direct Flight: Rethinking Audio-Guided Image Segmentation
    
}

\usepackage[colorlinks=true,linkcolor=blue,citecolor=myblue,urlcolor=black]{hyperref}
\hypersetup{hypertexnames=false} 

\author{Joel A. Santos$^{1}$, Zongwei Wu$^{1}$, Xavier Alameda-Pineda$^{2}$, Radu Timofte$^{1}$%
\\ [0.3em]  \normalsize \href{https://github.com/joelsantosvalle/AGIS}{https://github.com/joelsantosvalle/AGIS} \\ [-1.56em]
\thanks{\raggedright\scriptsize $^{1}$ Computer Vision Lab, CAIDAS \& IFI, University of Würzburg, Germany. }
\thanks{\raggedright\scriptsize $^{2}$ Inria at Univ. Grenoble Alpes, CNRS, LJK, France}%
}

\begin{document}

\tcbset{
  lavenderbox/.style={
    colback=lavenderbg,
    colframe=lavenderborder,
    arc=2mm,
    boxrule=0.5pt,
    left=2mm, right=2mm, top=1mm, bottom=1mm,
  }
}

\maketitle
\vspace{-5.5em}  

\begin{abstract}
Understanding human instructions is essential for enabling smooth human-robot interaction. In this work, we focus on object grounding, i.e., localizing an object of interest in a visual scene (e.g., an image) based on verbal human instructions. Despite recent progress, a dominant research trend relies on using text as an intermediate representation. These approaches typically transcribe speech to text, extract relevant object keywords, and perform grounding using models pretrained on large text-vision datasets. However, we question both the efficiency and robustness of such transcription-based pipelines. Specifically, we ask: Can we achieve direct audio-visual alignment without relying on text? To explore this possibility, we simplify the task by focusing on grounding from single-word spoken instructions. We introduce a new audio-based grounding dataset that covers a wide variety of objects and diverse human accents. We then adapt and benchmark several models from the closely audio-visual field. Our results demonstrate that direct grounding from audio is not only feasible but, in some cases, even outperforms transcription-based methods, especially in terms of robustness to linguistic variability. Our findings encourage a renewed interest in direct audio grounding and pave the way for more robust and efficient multimodal understanding systems.
\end{abstract}

\vspace{1ex}

\section{INTRODUCTION}

As human–robot interaction becomes more common in everyday settings, there is a growing need of methods that are natural, intuitive, and efficient in dynamic environments.  Since humans predominantly communicate through speech, recent work \textcolor{blue}{\cite{c01, c02}} has focused on enabling robots to understand verbal prompts for more adaptive behaviors. Most existing pipelines, however, rely on text as an intermediate representation: speech is transcribed via automatic speech recognition (ASR) and then used for visual reasoning \textcolor{blue}{\cite{c01}}. This sequential design suffers from two drawbacks: (i) compounded errors from imperfect transcription and language understanding \textcolor{blue}{\cite{c43 ,c004, c0004}}, and (ii) increased complexity and latency from chaining multiple modules \textcolor{blue}{\cite{c005}}. While advances in audio robustness \textcolor{blue}{\cite{c28, c29}}, transcription accuracy \textcolor{blue}{\cite{c36, c37}}, intent recognition \textcolor{blue}{\cite{c10}}, and text–vision alignment \textcolor{blue}{\cite{c34, c38, c39, c40}} mitigate some issues, the approach remains fragmented and text-centric, unlike human verbal communication which does not naturally pass through written language.

In this work, we ask: \textit{can spoken language be directly grounded to visual content without transcription?} While there exist some prior studies \textcolor{blue}{\cite{c07, c16, c17}} in audio-visual alignment, they typically focus on tasks such as localizing sound-producing objects, which differ in objective from ours. Here, we study object grounding from spoken referring expressions. To this end, we construct a dataset with clean, noise-free audio in a keyword-only setting. This is motivated by prior work \textcolor{blue}{\cite{c10}} showing that intent in referring tasks is often localized in a small set of salient words. Typically, in our dataset, each keyword corresponds directly to an object label (e.g., “mouse,” “keyboard", “tennis racket"), reducing ambiguity in sentences so to solely focus on the grounding and alignment part.

\begin{figure}[t]
\centering
\vspace{1mm}
\begin{overpic}[width=\linewidth]{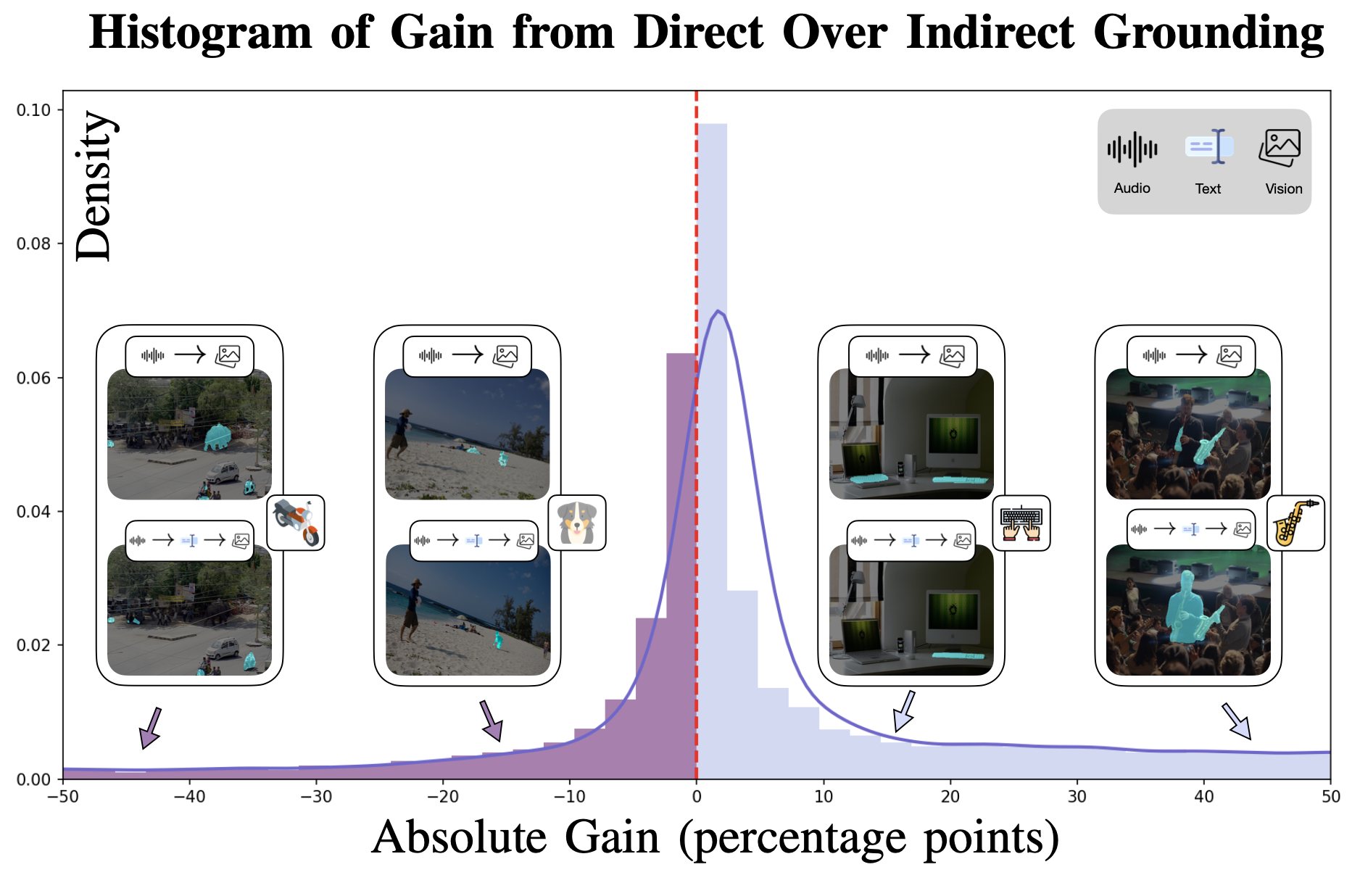}
  
  
\end{overpic}
\caption{\textbf{Histogram on the gain.} Direct speech grounding outperforms transcription grounding based on 5881 of 9905 images ($\approx $ 60\%). The models used for comparison correspond to the best performing representatives of their respective categories. While image grounding has traditionally been approached as a text-driven task, achieving remarkable success in recent years \textcolor{blue}{\cite{c34, c38, c39}}. When combined with ASR for natural human–computer interaction, the pipeline can accumulate errors. In this work, we explore direct speech-driven grounding, bypassing intermediate text conversion, which improves robustness to subtle tonal variations often overlooked by text-based models.}  
\label{fig:teaser}
\vspace{-4mm}
\end{figure}

Although one may expect that chaining audio-to-text and text-to-image models would yield perfect grounding, our benchmark reveals that this approach is far from being satisfactory, even under this oracle and simplest setting. Figure \ref{fig:teaser} illustrates the histogram of gains, comparing the top performing models in their respective categories. The results highlight a major challenge: ambiguity in both spoken language (even when simplified to keywords) and the visual domain. Text-based pipelines remain sensitive to accents, intonation, and subtle prosodic cues, introducing omissions and misinterpretations. Lexical ambiguity and imprecise punctuation further harm accuracy. Differently, we find that direct speech-to-vision grounding avoids these issues by learning directly from raw audio signals, thereby removing dependence on fragile transcription and enabling a more robust multimodal alignment.

A second challenge emerges in multi-object scenes: text-based models tend to produce more precise, conservative segmentations, while speech-driven models segment more inclusively but with slightly per-object lower precision. This may reflect differences in inductive biases, i.e., text models benefit from large-scale, curated pretraining and linguistic conventions, whereas speech models may exploit broader contextual or co-occurrence cues. This discrepancy reveals a fundamental modality-specific trade-off, underscoring the need to better understand and leverage these differences which are the central focus of this paper.

To summarize, we contribute to a foundation for evaluating the robustness of object grounding algorithms driven by audio inputs in complex visual scenes. We introduce a new benchmark that systematically compares transcription-based and direct speech-to-vision grounding pipelines, providing in-depth insights regarding the cross-modal alignment solely. Our findings highlight modality-specific differences in handling linguistic ambiguity and underscore the potential of direct speech-driven approaches as a compelling alternative to traditional text-based methods.

\section{RELATED WORK}

\textbf{Referring Image Segmentation (RIS)} is the task of segmenting a specific object in an image based on a natural language expression that uniquely identifies it. Recent work in this field has proposed increasingly sophisticated models to bridge visual and linguistic understanding \textcolor{blue}{\cite{c34, c38}}. \textcolor{blue}{\cite{c11}} introduced a two-stage pipeline: Location, then Segment, which first localizes the target region using the referring expression and then performs precise segmentation within that region. This separation of localization and segmentation allows for more accurate and interpretable outputs, especially in complex scenes. Complementing this, \textcolor{blue}{\cite{c12}} proposed the Encoder Fusion Network, which uses a co-attention mechanism to jointly embed and align visual and textual features. Their approach enhances the interaction of features between the modalities, resulting in better segmentation performance by enabling the model to focus on the semantically relevant regions of the image. \textbf{Referring Audio-Visual Segmentation (Ref-AVS)} \textcolor{blue}{\cite{c09}} extends RIS by using audios, text expression and visual information to identify target objects. This reflects more realistic human-robot interaction scenarios, where instructions are typically given verbally and may include hesitations, informal phrasing, or background noise.  

 \textbf{Sound source localization (SSL)} is a fundamental task in multimodal learning that aims to identify the spatial origin of a sound source within a visual scene. Traditional approaches typically associate visual regions with the audio signal, assuming that objects emitting sound can be localized through visual correspondence. Recent advancements have significantly improved localization accuracy by focusing on better cross-modal representations. For example, \textcolor{blue}{\cite{c06}} highlights that effective SSL relies heavily on cross-modal alignment between auditory and visual cues, emphasizing the importance of learning robust joint embeddings. \textcolor{blue}{\cite{c41}} further refines this process by proposing a method to reduce false negatives during localization, thereby enhancing precision in complex scenes. Moreover, methods such as T-VSL by \textcolor{blue}{\cite{c42}} incorporates textual guidance to improve localization in sound mixtures, extending SSL into the realm of tri-modal reasoning (audio, vision, and language). Similarly, \textcolor{blue}{\cite{c14}} proposes an approach to localize sound sources in mixtures without prior source knowledge, making SSL more adaptable to real-world, unstructured environments. 

\textbf{Audio visual segmentation (AVS)} is a multimodal task that aims to segment objects in visual scenes based on associated audio signals, enabling a deeper understanding of events where both appearance and sound provide complementary cues \textcolor{blue}{\cite{c16, c18, c19, c20, c23, c24, c25, c26, c27}}. The field has seen rapid progress with the introduction of diverse models that explore different aspects of cross-modal fusion. Early work by \textcolor{blue}{\cite{c07}} laid the foundation for AVS by establishing the basic task of identifying and segmenting sounding objects using both visual and auditory cues. Subsequent efforts, such as TransAVS \textcolor{blue}{\cite{c16}} and AVSegFormer \textcolor{blue}{\cite{c24}}, introduced transformer-based architectures to better capture long-range dependencies across modalities. Other methods have explored instance-level reasoning \textcolor{blue}{\cite{c17}}, motion-aware segmentation \textcolor{blue}{\cite{c20}}, and prompt-based strategies for class-aware segmentation \textcolor{blue}{\cite{c22}}. Several approaches have also addressed practical challenges, such as handling unlabeled frames \textcolor{blue}{\cite{c19}}, annotation-free learning \textcolor{blue}{\cite{c26}}, and improving generalization through bidirectional generation \textcolor{blue}{\cite{c27}}. Variational autoencoders \textcolor{blue}{\cite{c25, c18}} and cooperative learning mechanisms \textcolor{blue}{\cite{c21}} further demonstrate the variety of perspectives being explored.

\begin{figure}[!t]
  \centering
  \vspace{4mm}
  \includegraphics[width=\linewidth]{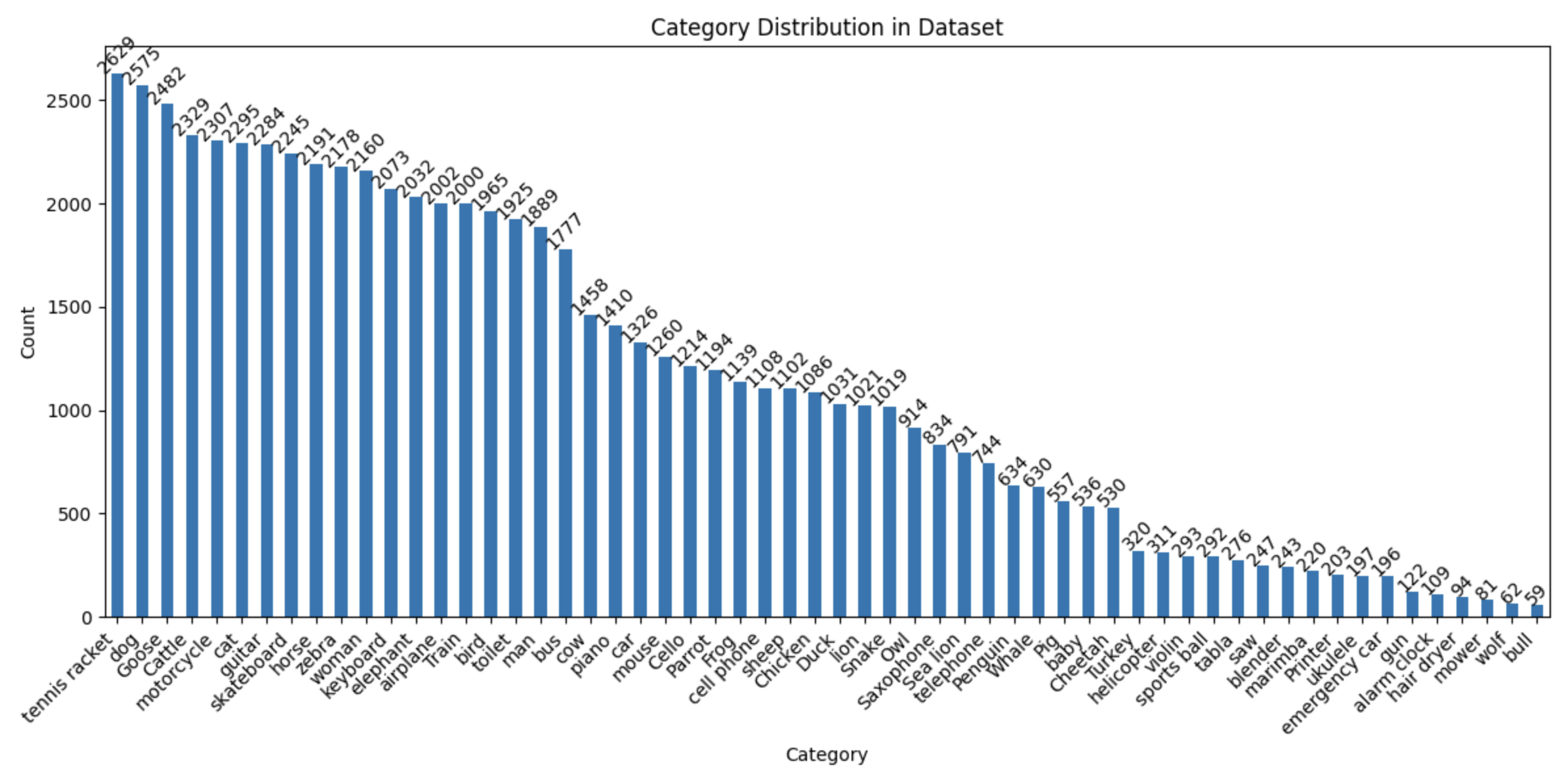}
  \vspace{-5mm}

  \caption{\textbf{Visual dataset distribution} composed of frames selected from the videos introduced in \textcolor{blue}{\cite{c07,c17,c26}}.}
\vspace{-4mm}
  \label{fig:dataset}

\end{figure}

\section{DATASET}


Despite significant advances in multimodal (Audio and Visual information) scene understanding \textcolor{blue}{\cite{c43}}, most existing methods \textcolor{blue}{\cite{c44}} still depend on textual instructions, an approach that fails to capture the rich, spoken interactions typical of real-world human–robot collaboration. Audio-driven techniques have begun to address this gap, but they generally concentrate on tasks such as background noise suppression or extracting keywords from extended sentences rather than on directly grounding spoken cues in visual data. In this work, we target precisely that challenge. To this end, we introduce a new dataset in which each image is paired with a perfectly generated audio, short and clean expression that refers to specific objects in the scene, recorded in a variety of English accents. This resource enables, for the first time, a systematic study of how feasible binding spoken references directly to their pixel-level counterparts.

\begin{table*}[ht!]
\centering  

  \vspace{4mm}
  \resizebox{\textwidth}{!}{\begin{tabular}{c | c|c|c|c|c|c|c|c} 
    \toprule
    \rowcolor{  gray!20} 
    Venue & Model & ASR model & Visual Backbone & F1-score & mIoU & GFlops  & Parameters & Inference time\\
    \midrule
    \multicolumn{9}{l}{\textbf{Audio-visual segmentation methods}}\\
    \midrule
     ECCV '22 & TPAVI \textcolor{blue}{\cite{c07}} & \begin{tabular}[c]{@{}c@{}}--\end{tabular} & Pvt & 69.4 & 54.2  & 66.2 & 101.3M& 16.6ms  \\
         ACM MM '23 & CATR \textcolor{blue}{\cite{c30}} & \begin{tabular}[c]{@{}c@{}}--\end{tabular} & Pvt & 72.2 & 55.1 & 263.2& 118.5M & 53.7ms  \\ 
    AAAI '24& AVSegformer \textcolor{blue}{\cite{c24}} & \begin{tabular}[c]{@{}c@{}}--\end{tabular} & Pvt & 63.7 & 51.3 & 94.1& 32.4M & 10.1ms \\
    ArXiv '24& PCMANET \textcolor{blue}{\cite{c32}} & \begin{tabular}[c]{@{}c@{}}--\end{tabular} & Pvt & 75.1 & 60.4 & 43.5 & 95.1M & 17.53ms \\

        CVPR '24& COMBO \textcolor{blue}{\cite{c21}} & \begin{tabular}[c]{@{}c@{}}--\end{tabular} & Pvt & 78.9 &  64.3 & 45.6 & 119.8M& 34.4ms \\
        ECCV '24& AAVS \textcolor{blue}{\cite{c31}} & \begin{tabular}[c]{@{}c@{}}--\end{tabular} & Swin-B & 82.3 & 69.5  & 145.1& 114.5M & 36.62ms \\

    \midrule
    \multicolumn{9}{l}{\textbf{Referring image segmentation + Automatic speech recognition methods}}\\
    \midrule


    
    &   & Whisper-base\textcolor{blue}{\cite{c36}} & Mamba-B & 73.9 & 61.6  & 283.7 * & 355.5M & 333.8ms \\
    ECCV '24& ReMamber \textcolor{blue}{\cite{c26}} & WavLM\textcolor{blue}{\cite{c37}} & Mamba-B & 68.9 & 57.2 & 283.7 *& 376M & 87.2 ms \\
    &   & Whisper-small\textcolor{blue}{\cite{c36}} & Mamba-B & 75.3 & 62.2 & 283.7 * & 525M  & 740.6 ms \\
    

    
    \midrule
    &   & Whisper-base\textcolor{blue}{\cite{c36}} & ViT-B & 59.2 & 46.3 & 16.94 * & 223.6M & 297.9 ms \\
    CVPR '25& HybridGL \textcolor{blue}{\cite{c38}} & WavLM\textcolor{blue}{\cite{c37}} & ViT-B & 55.6 & 42.1 &16.94 * & 244.6M& 55.3 ms\\
    &  & Whisper-small \textcolor{blue}{\cite{c36}} & ViT-B & 60.2 & 48.1 & 16.94 * & 339M& 708.8 ms \\

    \bottomrule
  \end{tabular}}

    \caption{\textbf{Quantitative results.} Performance comparison of models is conducted across direct speech-to-vision segmentation and transcription-based methods using our dataset. Values marked with  * indicate that the measurement is taken only from the RIS model.}
  \label{tab:commands}
\end{table*}

\textbf{Visual Images} The visual data was created by combining multiple existing datasets, specifically \textcolor{blue}{\cite{c07}, \textcolor{blue}{\cite{c17}}}, and \textcolor{blue}{\cite{c26}}. These original datasets were designed for video-based tasks; however, in our case, we simplify by extracting a single frame from each video, typically the first frame. The rationale behind this choice is that, once the object is localized in the first frame, video segmentation methods can then be applied to ensure temporally consistent tracking throughout the entire video. The resulting image set includes both synthetic and real-world images. After merging all sources, we obtained 66,202 images to pair with our audio samples. Figure \textcolor{green}{\ref{fig:dataset}} shows the category distribution of the visual dataset. 

\textbf{Synthesized Speech} The audio part consists of more than 2,300 samples covering 35 different accent types. All audio was synthetically generated using tools such as Google Text-to-Speech (gTTS), gTTS Cloud, and Amazon Polly. The accents include American, British, and other variations of native English speakers. The use of diverse English accents in the synthetic audio led to a range of patterns, with noticeable differences in how commands were perceived. For example, male voices often conveyed a more imperative tone compared to female voices, despite identical content. These subtleties were observed to affect downstream tasks, particularly during transcription. A single keyword, such as ``keyboard,” could be transcribed differently depending on the speaker’s emphasis, appearing as ``keyboard!” when spoken with intensity or as ``keyboard.” when said more neutrally. These differences, shaped by both intonation and accent, introduced subtle but meaningful variation in the transcriptions, which could influence the performance of models relying on text inputs.

\section{BENCHMARK}

To evaluate the alignment between image and audio modalities, we introduce a novel benchmark for direct audio-guided image segmentation. We first adapt leading methods from the AVS tasks, including TPAVI \textcolor{blue}{\cite{c07}}, CATR \textcolor{blue}{\cite{c30}}, AVSegformer, \textcolor{blue}{\cite{c24}}, PCMANET \textcolor{blue}{\cite{c32}}, COMBO \textcolor{blue}{\cite{c21}}, and AAVS \textcolor{blue}{\cite{c31}}. Technically, we modify their original video-based and temporal setting by replacing video streams with single image frames and limiting audio input to one-second clips. We keep the training setting as the original works. We then perform a comprehensive evaluation by retraining these adapted AVS methods on our proposed benchmark. For comparison, we also implement a typical robotics pipeline (transcription model): multiple SOTA ASR model transcribe the audio, and the resulting text is passed to a RIS model. In this baseline, we use the original publicly released checkpoints for both ASR and RIS, as they were trained on datasets substantially larger than ours (with 388,334 expressions across more than 50,000 objects \textcolor{blue}{\cite{c34}}). To assess the quality of the predicted outputs, we use the mIoU and the F1 score.

\section{Quantitative Comparison}

Table \textcolor{blue}{\ref{tab:commands}} shows the quantitative performances. 
We can observe that the results across both direct and transcription based pipelines reveal that higher computational cost does not necessarily lead to better performance. Within the direct category, certain models achieve the highest accuracy while maintaining moderate size and efficient inference, outperforming others that are larger and more computationally demanding. These findings show that model size and raw FLOPs are poor indicators of real-world effectiveness, and that efficient architectures with well-designed fusion mechanisms can consistently deliver superior accuracy at a fraction of the resource cost. 

This contrast is even more evident when comparing transcription models to the proposed direct framework. Transcription models approaches require running both a speech recognition model and a separate segmentation backbone, resulting in massive parameter counts, high computational load, and slower inference. By contrast, direct audio-visual segmentation methods maintain a streamlined architecture with equal or superior accuracy, while eliminating the intermediate transcription step. This integrated design not only reduces computational requirements and inference time but also avoids bottlenecks that can degrade performance, making it far better suited to real-time robotics applications where efficiency and rapid responsiveness are critical.

\section{Qualitative Comparison}

\subsection{Speech-Vision Grounding}

\begin{figure}[t]
    \centering

\begin{tcolorbox}[
        enhanced,
  boxsep=1mm, top=0mm, bottom=0mm, left=0mm, right=0mm,
    overlay={
  },
]
        \begin{overpic}[width=1.0\textwidth]{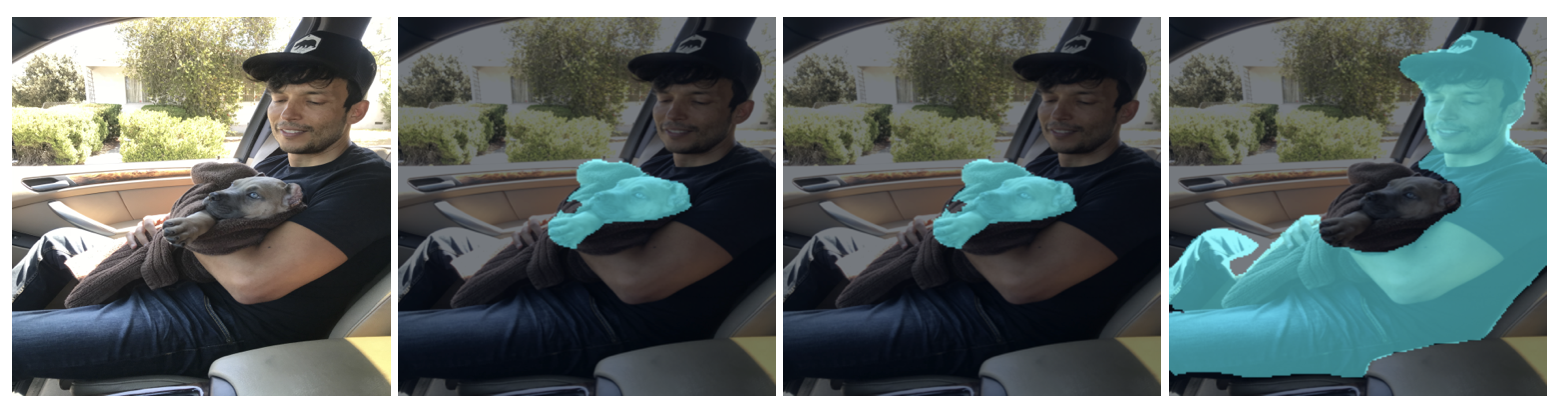}
        \put(24.4,20.5){
          \begin{tikzpicture}
            \node[
              draw,            
              fill=white,      
              rounded corners=1pt,
              inner sep=2pt
            ] {\includegraphics[width=4mm]{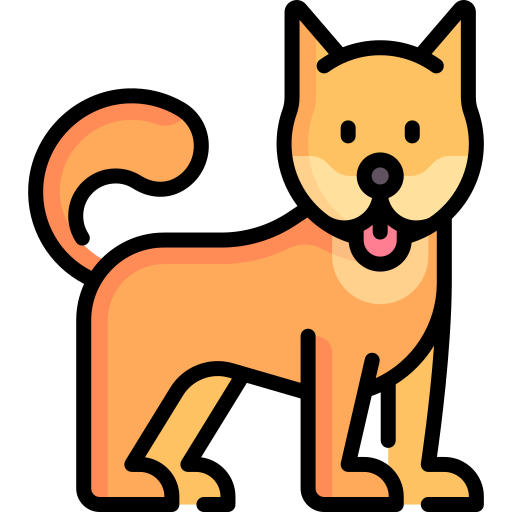}};
          \end{tikzpicture}
        }
        \put(49,20.5){
          \begin{tikzpicture}
            \node[
              draw,            
              fill=white,      
              rounded corners=1pt,
              inner sep=2pt
            ] {\includegraphics[width=4mm]{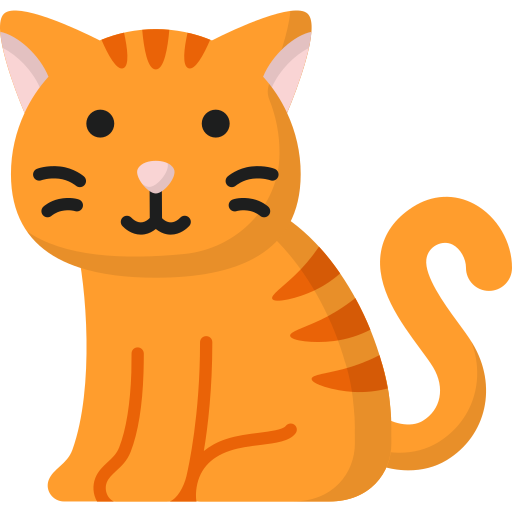}};
          \end{tikzpicture}
        }

        \put(73.7,20.5){
          \begin{tikzpicture}
            \node[
              draw,            
              fill=white,      
              rounded corners=1pt,
              inner sep=2pt
            ] {\includegraphics[width=4mm]{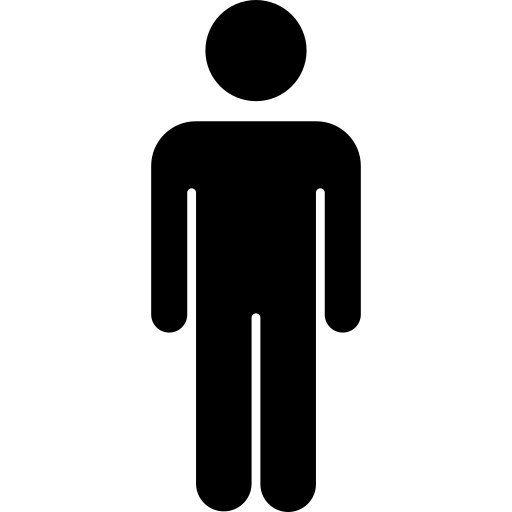}};
          \end{tikzpicture}
        }
        \end{overpic}
    \end{tcolorbox}

\begin{tcolorbox}[
        enhanced,
  boxsep=1mm, top=0mm, bottom=0mm, left=0mm, right=0mm,
    overlay={
  },
]
        \begin{overpic}[width=1.0\textwidth]{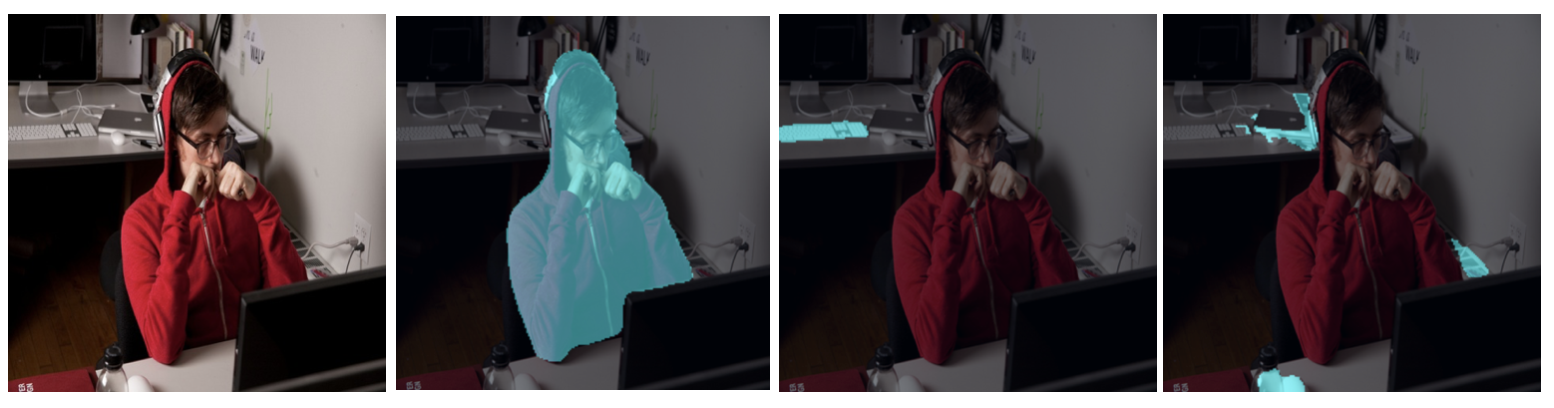}
        \put(24.4,21.5){
          \begin{tikzpicture}
            \node[
              draw,            
              fill=white,      
              rounded corners=1pt,
              inner sep=2pt
            ] {\includegraphics[width=4mm]{images/man.png}};
          \end{tikzpicture}
        }
        \put(49,21.5){
          \begin{tikzpicture}
            \node[
              draw,            
              fill=white,      
              rounded corners=1pt,
              inner sep=2pt
            ] {\includegraphics[width=4mm]{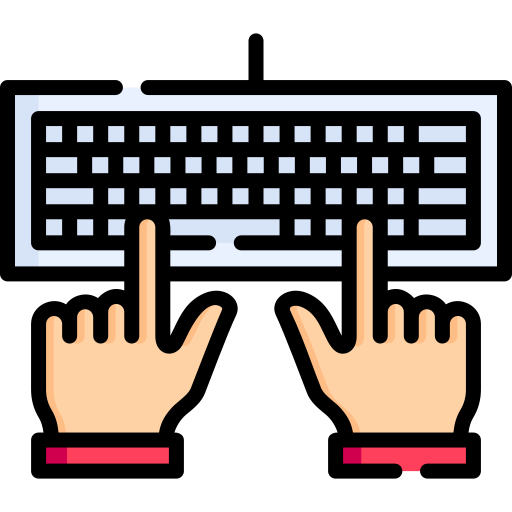}};
          \end{tikzpicture}
        }

        \put(73.7,21.5){
          \begin{tikzpicture}
            \node[
              draw,            
              fill=white,      
              rounded corners=1pt,
              inner sep=2pt
            ] {\includegraphics[width=4mm]{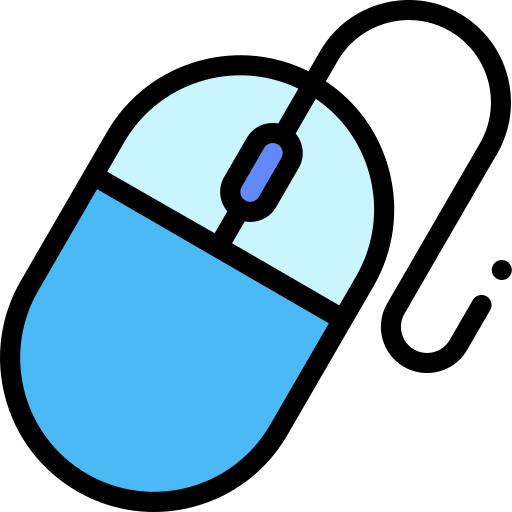}};
          \end{tikzpicture}
        }
        \end{overpic}
    \end{tcolorbox}

\begin{tcolorbox}[
  enhanced,
  boxsep=1mm, top=0mm, bottom=0mm, left=0mm, right=0mm, colback=white,
  overlay={
  },
]
\centering

\begin{overpic}[percent,width=\linewidth]{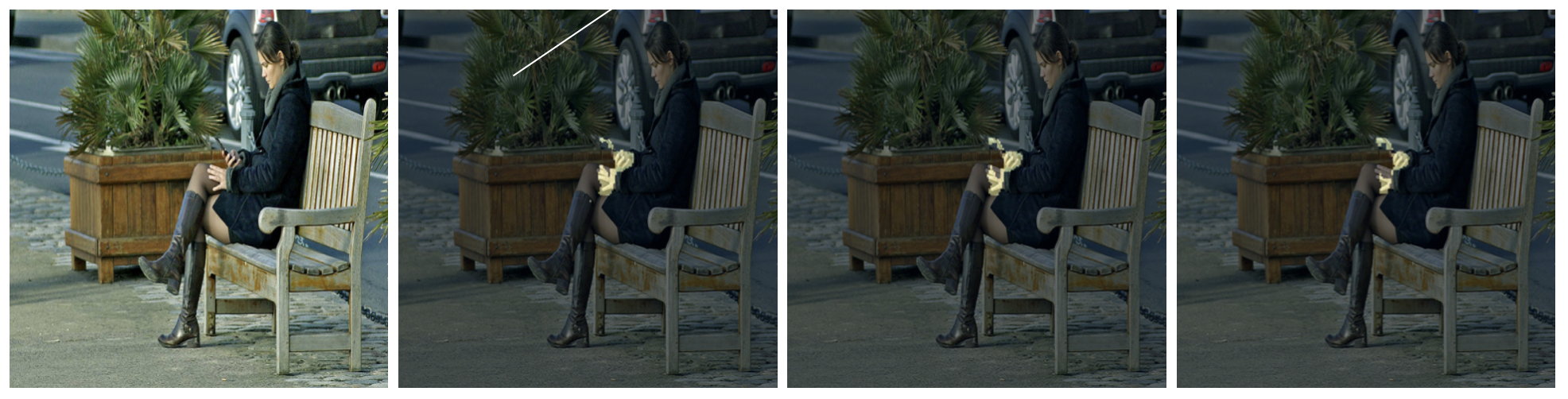}
  \put(24.4,21.5){%
    \begin{tikzpicture}
      \node[draw,fill=white,rounded corners=1pt,inner sep=2pt]
        {\includegraphics[width=4mm]{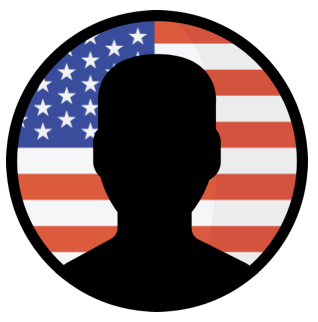}};
    \end{tikzpicture}}
  \put(49.2,21.5){%
    \begin{tikzpicture}
      \node[draw,fill=white,rounded corners=1pt,inner sep=2pt]
        {\includegraphics[width=4mm]{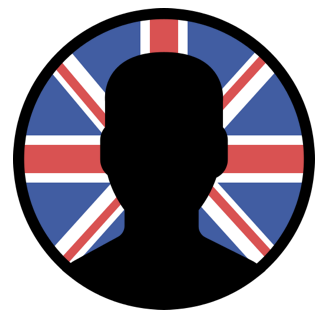}};
    \end{tikzpicture}}
  \put(74.1,21.5){%
    \begin{tikzpicture}
      \node[draw,fill=white,rounded corners=1pt,inner sep=2pt]
        {\includegraphics[width=4mm]{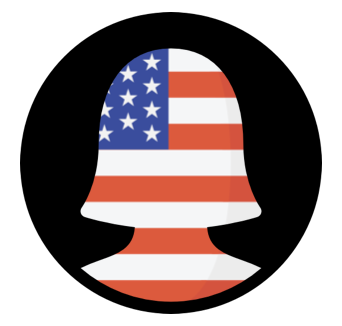}};
    \end{tikzpicture}}
\end{overpic}

\begin{overpic}[percent,width=\linewidth]{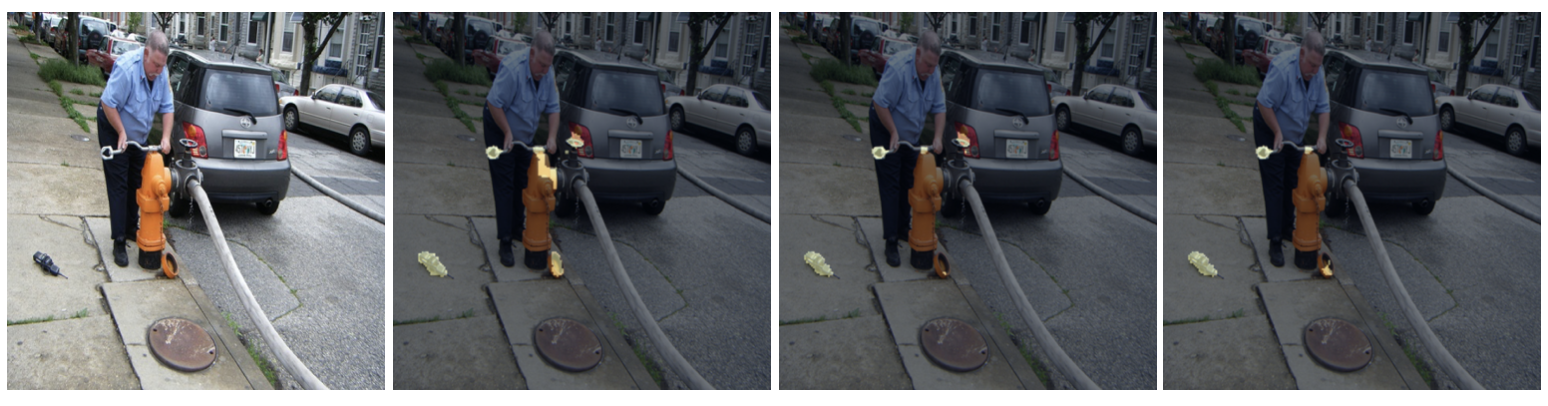}

\end{overpic}

\end{tcolorbox}
\vspace{-2.5mm}

    \caption{\textbf{Qualitative results from our benchmark.} Top rows: multi-object cases. Bottom row: accent variations.}
    \label{fig:enter-label}
    \vspace{-4mm}

\end{figure}

Figure \textcolor{blue}{\ref{fig:enter-label}} illustrates model behavior across different test cases, focusing on semantic nuances and accent effects in audio inputs. In the first row, varying the keyword reveals semantic ambiguity: dog yields correct segmentation, while cat, absent in the scene, still segments the dog, suggesting confusion between visually similar categories. Meanwhile, man is segmented accurately. 

The second row shows multi-object scenes. Keywords man and keyboard produce precise segmentations, while mouse results in minor degradation but still selects the object closest to the user.

The last two columns assess accent robustness: heavier accents, such as a male accent, increase false positives and segmentation drift, whereas lighter accents yield cleaner results. Overall, performance is sensitive to both input semantics and audio clarity, with ambiguous cues often triggering confusion between similar objects.

\begin{figure*}[!t]
  \centering

\includegraphics[width=\linewidth]{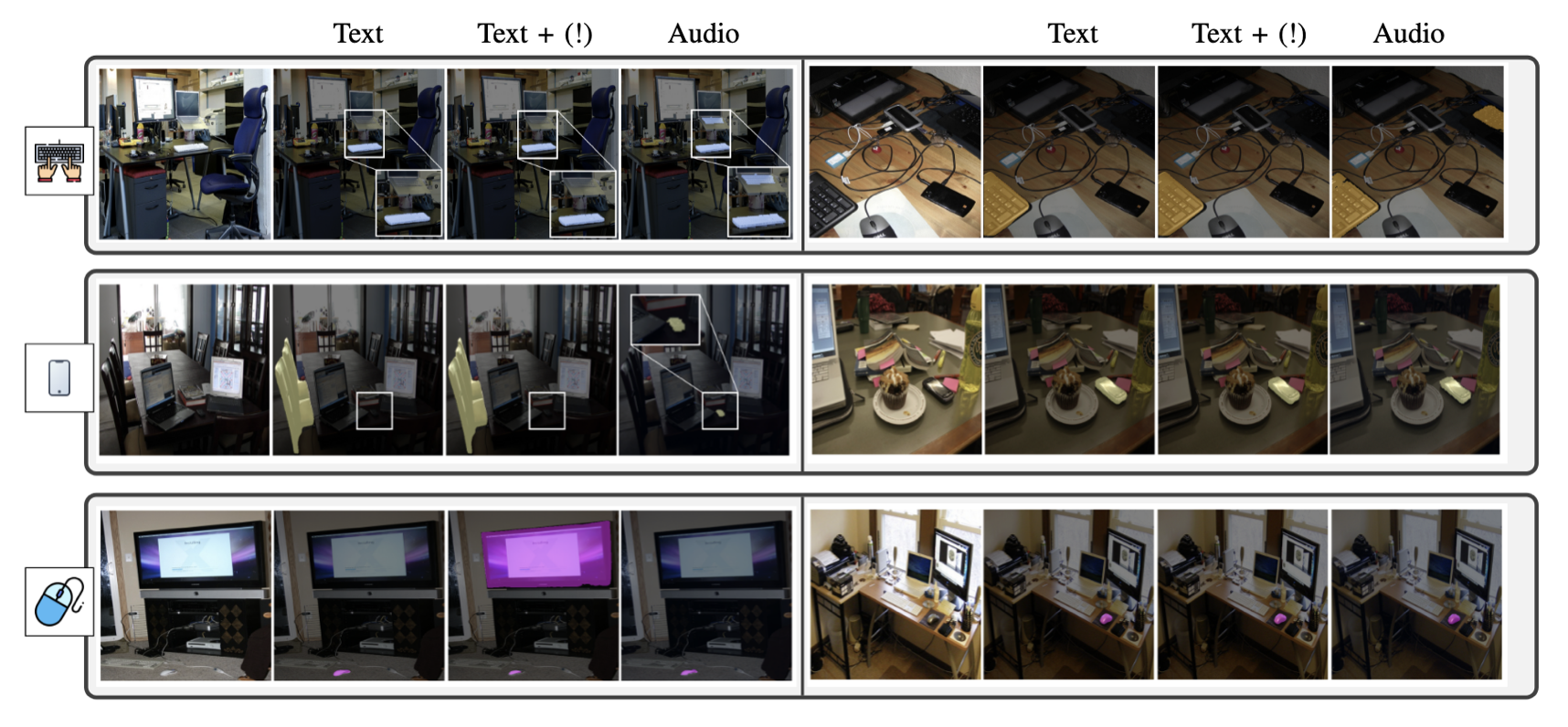}

\vspace{-2.5mm}

  \caption{\textbf{Qualitative Results.} Segmentation results under varied conditions, including multiple object classes, multiple instances of the same class, and single objects with phrasing variations. The middle columns show transcription model outputs with different intonations captured, while the final column presents outputs from the direct speech to vision system.}
  \label{fig:fullstack}
  \vspace{-3mm}

\end{figure*}

\begin{figure*}[!t]
  \centering

    \includegraphics[width=\linewidth]{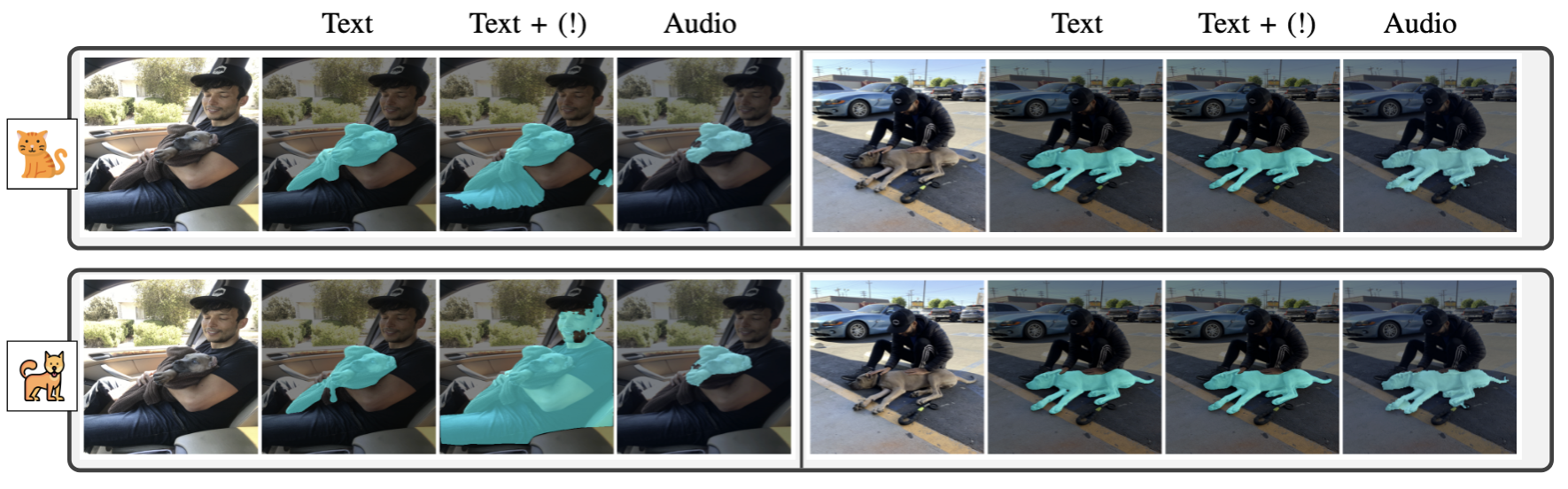}
\vspace{-5.5mm}

  \caption{\textbf{Qualitative Results.} Discrepancies arise between the input text or audio descriptions and the corresponding visual scenes, particularly when objects that are visually similar to each other are present in the visual scene. The performance of transcription models is notably and substantially reduced due to errors in interpretation, such as the misidentification of ``sea lion" as the phrase ``See you, Lion!" This type of mistake leads to significant inaccuracies in the segmentation process}
  \label{fig:fullstack_two}
  \vspace{-3mm}

\end{figure*}

\subsection{Text- and Speech-Vision Grounding}

In the first two rows of Figure \textcolor{blue}{\ref{fig:fullstack}}, each row shows scenes where the target object appears multiple times. For example, the first row contains two distinct keyboards. Although our transcription system occasionally inserted punctuation marks (e.g., period) based on speech tone, these had minimal effect on segmentation quality. The transcription model, consistently segmented only a single instance of the object class, even when multiple instances were clearly visible. In contrast, a direct audio-vision method accurately segmented all instances without error. Even when the transcription was explicitly modified to request both keyboards, the transcription models still failed to detect more than one instance, underscoring a limitation in their grounding capability.

The third and fourth row of the Figure \textcolor{blue}{\ref{fig:fullstack}} highlights cases involving smaller, and occluded objects. In some examples, the transcription models manage to segment the correct object but in other the models either missed the correct object entirely or produced spurious masks over irrelevant areas. For example, in the last row, while the transcription model did identify the mouse, it also incorrectly highlighted nearby regions, leading to an imprecise result. In contrast, the direct audio-vision model successfully isolated only the intended object. This robustness stems from the fact that our system directly processes raw audio, avoiding dependency on transcribed punctuation or sentence structure. As a result, variations in speaker accent, intonation, or automatic punctuation, common sources of noise in transcription based pipelines, have little to no effect on the direct approach.

Figure \textcolor{blue}{\ref{fig:fullstack_two}} illustrate cases where mismatches occur between the input text or audio and the visual scene, particularly when a visually similar animal is present. For example, in one case both images contained a dog, yet when the target keyword was cat, the model still segmented the dog, indicating a semantic confusion between similar object categories. In some instances, the addition of intonation cues led to degraded mask quality, and the tokenization of punctuation (e.g., “!”) appeared to distract the model, reducing segmentation accuracy. These effects are amplified in transcription-based pipelines, where recognition errors compound the problem: for example, the phrase “cellphone” was transcribed as “SILFUN”, leading the model to segment irrelevant regions. Such cases illustrate how cross-modal grounding is vulnerable not only to visual similarity, but also to imperfections in the linguistic channel itself, resulting in degraded segmentation quality or the inclusion of unrelated objects.

\section{Ablation Studies and Discussions}

\begin{figure}[!t]
  \centering
  \vspace{2mm}
  \includegraphics[width=\linewidth]{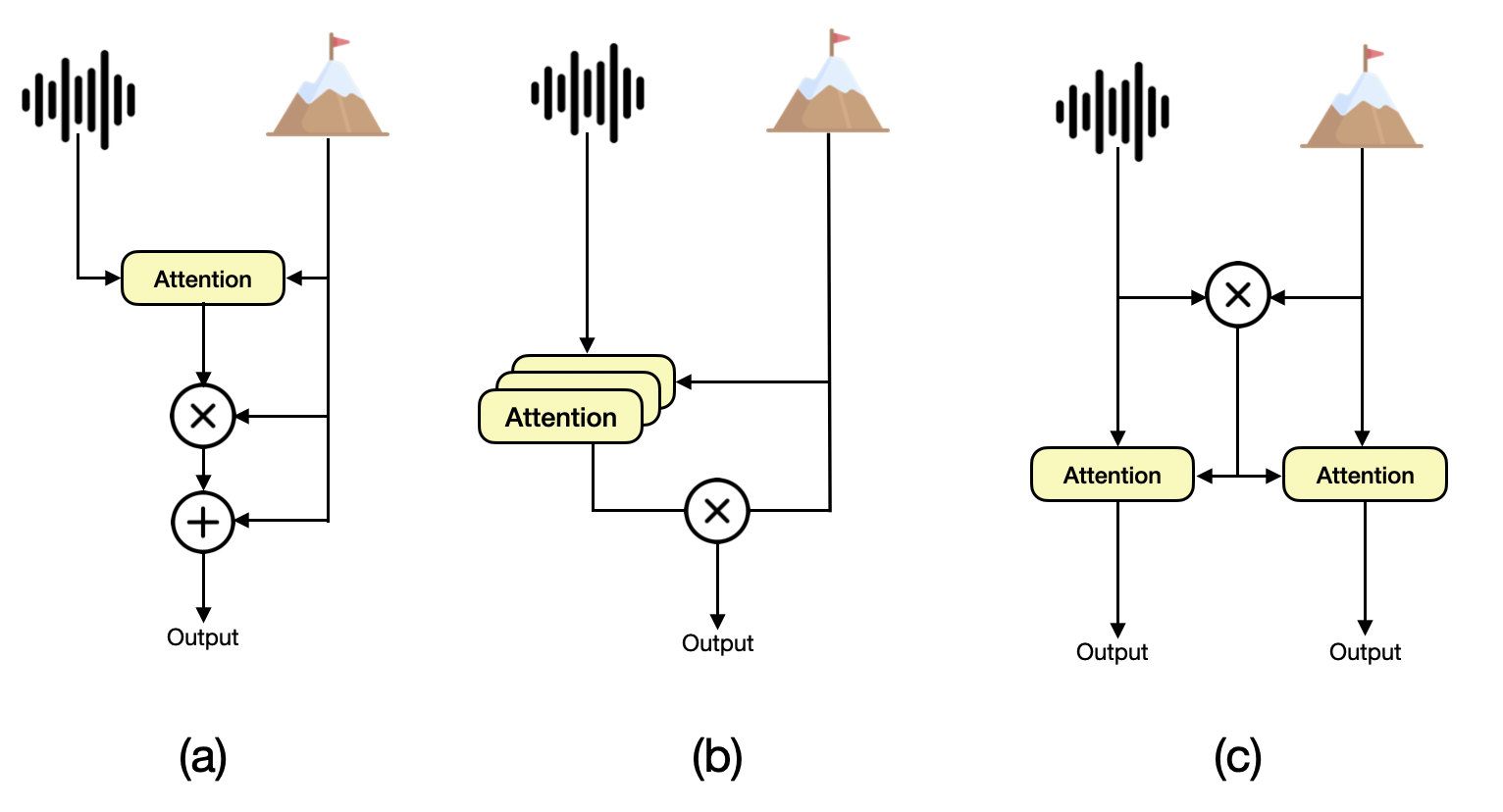}
  \caption{\textbf{Architecture of the different modality fusion modules employed by the AVS models.} The subfigures represent the models as follows: (a) Cross-modal fusion \textcolor{blue}{\cite{c07}}, (b) Group-wise fusion \textcolor{blue}{\cite{c31}}, and (c) Bilateral alignment \textcolor{blue}{\cite{c21}}.} 
  \label{fig:four_nolabels}

\end{figure}

In the literature, alignment methods can be broadly categorized into three main types: bilateral alignment, cross‐modal fusion, and group‐wise fusion, as exemplified by COMBO \textcolor{blue}{\cite{c21}}, TPAVI \textcolor{blue}{\cite{c07}}, and PCMANET \textcolor{blue}{\cite{c32}}, respectively. An overview of these fusion pipelines is provided in Figure \textcolor{blue}{\ref{fig:four_nolabels}}. Additionally, we consider naive fusion among the analyzed mechanisms, where the audio embedding is directly added to the visual feature map and subsequently projected back through convolution. Since our primary focus is on cross‐modal alignment, we aim to carefully isolate and assess the effect of different fusion strategies on model performance. To achieve this, we conduct ablation experiments using COMBO \textcolor{blue}{\cite{c21}} as the baseline architecture, systematically substituting its original fusion module with alternative designs. COMBO \textcolor{blue}{\cite{c21}} was selected due to its optimal balance between segmentation accuracy and computational cost (GFLOPs), as well as its strong performance when combined with the most widely adopted backbone. All models are evaluated under strictly identical experimental conditions, ensuring a fair and reproducible comparison that attributes observed differences solely to the fusion mechanism itself.

Table \textcolor{blue}{\ref{tab:example}} presents the quantitative results. Notably, while different fusion designs introduce measurable variations in performance, the overall metric ranges remain relatively stable because the other network components, such as the encoder, decoder, and training configuration are held constant. This controlled setup isolates the influence of the fusion mechanism itself. As expected, less sophisticated fusion strategies exhibit a noticeable drop in quantitative performance, largely due to their reduced capacity to model fine-grained cross-modal correspondences and effectively align semantic information between modalities. In contrast, more advanced designs are better able to preserve subtle cross-modal cues, resulting in improved segmentation quality and more consistent predictions. 

\medskip


\begin{table}

\centering
\vspace{4mm}
\renewcommand{\arraystretch}{1.2} 
\setlength{\tabcolsep}{6pt} 
\begin{tabular}{c | c | c | c}
\toprule
\rowcolor{gray!20} 
Model & Fusion & F1-score & mIoU \\
\midrule
\multirow{4}{*}{COMBO \textcolor{blue}{\cite{c21}}} 
 & Bilateral  \textcolor{blue}{\cite{c21}}  & 78.9 & 64.7 \\
 & Cross-Modal \textcolor{blue}{\cite{c07}} & 79.5 & 65.1 \\
 & Group‐wise \textcolor{blue}{\cite{c31}}  & 79.2 & 65.4 \\
 & Naive        & 78.4 & 64.1 \\
\bottomrule
\end{tabular}

\caption{\textbf{Quantitative results.} Performance comparison of the selected model after swapping the fusion module.}
\label{tab:example}
\vspace{-4.85mm}

\end{table}

\begin{figure}[!t]

\centering

\vspace{3mm}

\includegraphics[width=\linewidth]{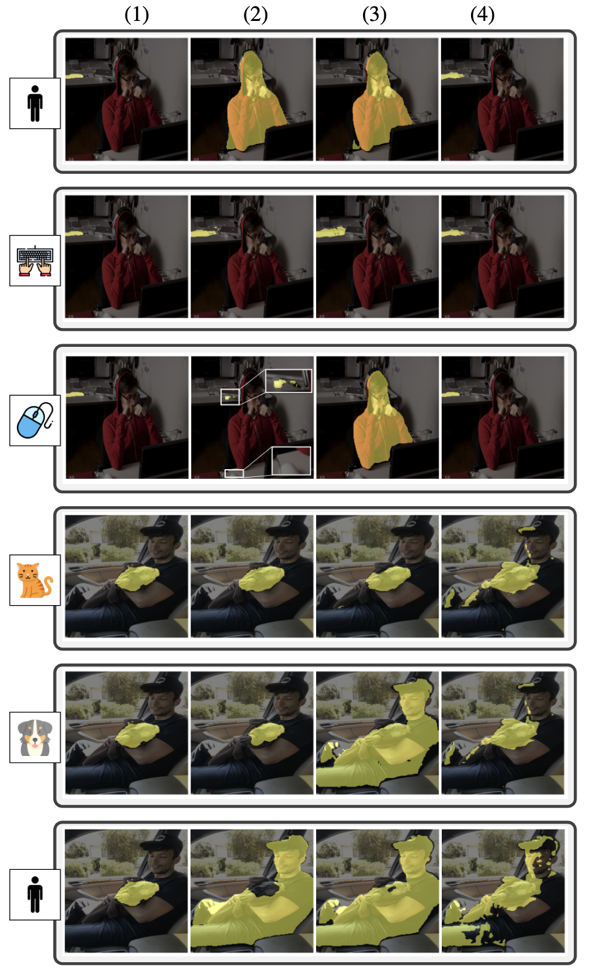}
\caption{\textbf{Qualitative results.} From left to right, the columns illustrate: (1) Bilateral fusion \textcolor{blue}{\cite{c21}}, (2) Cross modal fusion \textcolor{blue}{\cite{c07}}, (3) Group-wise fusion \textcolor{blue}{\cite{c31}}, and (4) Naive fusion where audio embeddings are directly added to the visual feature map, then projected back.}
\vspace{-4mm}
    \label{fig:last oneeee}
    
\end{figure}

In Figure \textcolor{blue}{\ref{fig:last oneeee}}, we present additional qualitative results. We further conduct an in-depth analysis, aligned with Figure \textcolor{blue}{\ref{fig:enter-label}}, by varying the audio intention for the same image. We can observe that Column 1 \textcolor{blue}{\cite{c21}} often ignores the audio guidance, selecting the same region or object regardless of the cue provided, suggesting an over-reliance on visual information at the expense of intention. By contrast, Column 2 \textcolor{blue}{\cite{c07}} can correctly localize all objects, even with occluded ones (e.g., ``mouse"). Column 3 \textcolor{blue}{\cite{c31}} performs well under straightforward conditions, as also reflected by its strong results in Table 2. However, in more challenging cases such as distinguishing occluded objects when other prominent classes are present, e.g., with ``mouse" or ``human" as speech prompt, its focus deteriorates. This suggests that its fusion strategy leverages global context but lacks sufficient class-specific grounding, leading to confusion under ambiguous speech-vision alignment. Column 4 with the naive fusion, as expected, shows the weakest performance since it employs the simplest strategy, failing to generalize and mainly identifying only the most prominent objects. 

We observe again the visual ambiguity that all models struggle to generalize when presented with four-legged animals; for example, audio cues for ``cat” and ``dog” often lead to the segmentation of whichever four-legged animal is visually present, regardless of the intended label. This behavior raises important questions about the true discriminatory capabilities: specifically, whether they can genuinely distinguish between cats and dogs, or if they predominantly rely on generic features common to quadrupeds. The high visual similarity between these animals likely causes the models to prioritize broad morphological cues over fine-grained semantic distinctions, suggesting limitations in the ability to resolve subtle category differences in cross-modal grounding.

\begin{tcolorbox}[lavenderbox]
\textit{\textbf{Takeaways:} (i) Different fusion modules yield varying degrees of cross-modal alignment, even when producing qualitatively similar outcomes.
(ii) While fusion design can partially alleviate ambiguity, fully addressing this challenge demands a deeper reconsideration of semantic disentanglement within feature representations. }
\end{tcolorbox}

\section{CONCLUSION}

In this work, we study the feasibility of direct speech to vision grounding and introduce a dedicated benchmark. Under a common evaluation setup, we adapt several AVS models and compare them to transcription based pipelines. Results show that direct grounding achieves higher accuracy with lower latency and parameter cost, yielding a simpler and more robust pipeline. Ablation studies reveal that fusion design is one of the primary drivers of cross-modal alignment: while quantitative results may appear similar, different modules yield qualitatively distinct behaviors and varying resilience to ambiguity in speech-vision alignment. Stronger fusion strategies can mitigate semantic confusion, though resolving fine-grained ambiguities likely requires more disentangled modality specific representations. We hope these findings will stimulate further research on direct speech-vision methods.

\phantomsection
\bibliographystyle{IEEEtran}
\bibliography{main_content/mybib}

\end{document}